\documentclass[lettersize,journal]{IEEEtran}
\usepackage{amsmath,amsfonts}
\usepackage{algorithmic}
\usepackage{algorithm}
\usepackage{array}
\usepackage[caption=false,font=normalsize,labelfont=sf,textfont=sf]{subfig}
\usepackage{textcomp}
\usepackage{stfloats}
\usepackage{url}
\usepackage{verbatim}
\usepackage{graphicx}
\usepackage{cite}
\usepackage{textcomp}
\usepackage{stfloats}
\usepackage{url}
\usepackage{verbatim}
\usepackage{graphicx}
\usepackage{amsmath,amssymb,amsfonts}
\usepackage{algorithmic}
\usepackage{graphicx}
\usepackage{textcomp}
\usepackage{xcolor}
\usepackage{stfloats}
\usepackage{lipsum}
\usepackage{amsfonts}
\usepackage{multirow}
\usepackage{booktabs}
\usepackage{makecell}
\usepackage{doi}
\usepackage{pifont}
\usepackage{etoolbox}
\begin{document}

\title{High-Fidelity and Controllable Face Editing via 3D-Aware Diffusion}

\author{Mengting Wei, 
      Tuomas Varanka,
      Yante Li,
      Xingxun Jiang,
      Huai-Qian Khor,
      Guoying Zhao$^{\ast}$,~\IEEEmembership{Fellow,~IEEE}

   \thanks{M. Wei, T. Varanka, H. Khor and G. Zhao are with the Center for Machine Vision and Signal Analysis, Faculty of Information Technology and Electrical Engineering, University of Oulu, Oulu, FI-90014, Finland. E-mail: mengting.wei@oulu.fi, varanka.tuomas@gmail.com, yante.li@oulu.fi, huai.khor@oulu.fi, guoying.zhao@oulu.fi.}
   \thanks{X. Jiang is with the Key Laboratory of Child Development and Learning Science of Ministry of Education, School of Biological Sciences and Medical Engineering, Southeast University, Nanjing 210096, China, and is also with the Center for Machine Vision and Signal Analysis, Faulty of Information Technology and Electrical Engineering, University of Oulu, Oulu, FI-90014, Finland (e-mail:jiangxingxun@seu.edu.cn).}
   \thanks{*Corresponding author}}

\markboth{Journal of \LaTeX\ Class Files,~Vol.~14, No.~8, August~2021}%
{Shell \MakeLowercase{\textit{et al.}}: A Sample Article Using IEEEtran.cls for IEEE Journals}

\IEEEpubid{0000--0000/00\$00.00~\copyright~2021 IEEE}

\maketitle

\begin{abstract}
Face editing involves modifying facial attributes like expression, head pose, or lighting, with the goal of preserving the subject’s unique identity features. Diffusion models have recently emerged as the dominant approach in visual generation, driven by their strong generative power. However, challenges persist in the realm of face editing, where realistically and accurately editing target attributes while preserving high-fidelity identity information remains a formidable problem. In this paper, we present RigFace, a novel framework that integrates 3DMM-based controllable signals with a fully fine-tuned Stable Diffusion (SD) model. We observe that this task essentially involves the combinations of target background, source identity and face attributes aimed to edit. Our goal is to leverage 3DMM to provide disentangled attribute control, and harness the strong generative capacity of SD to produce photorealistic face images. Specifically, RigFace contains: 1) A Spatial Attribute Encoder that provides presise and decoupled conditions of background, pose, expression and lighting; 2) A high-consistency FaceFusion module that transfers identity features from the Identity Encoder to the Denoising UNet of a pre-trained SD model through self-attention; 3) A full fine-tuning strategy that enables the SD model to better adapt to our task. Our model achieves superior performance in both identity preservation and photorealism compared to existing face editing models.
\end{abstract}


\section{Introduction}
\label{sec:intro}

Realistically editing facial expression, head pose, and lighting while preserving identity and high-frequency details is critical for numerous applications such as personalized virtual avatars, face reenactment, and film production \cite{tellamekala20233d,sun2023continuously,lee2014intra,11251121,amini2015hapfacs}. These tasks demand high photorealism and precise identity preservation, which remain challenging due to the complexity of disentangling and manipulating fine-grained facial attributes without introducing visual artifacts.

To achieve controllable editing of facial attributes, 3D Morphable Models (3DMMs) have been widely adopted due to their interpretable parameterization of expression, pose, and lighting \cite{deng2019accurate,hong2022headnerf,feng2021learning}. Off-the-shelf estimators such as DECA \cite{feng2021learning} allow extraction of these parameters from real-world images. However, directly rendering from 3DMM parameters typically leads to CGI-like appearances caused by oversimplified geometry, coarse reflectance models, and lack of fine details like hair and accessories. This has motivated hybrid approaches that combine 3DMM controls with data-driven generative models. A substantial body of prior work focuses on integrating 3DMM controls with GAN-based models \cite{bounareli2023hyperreenact,ma2025review,jiang2023talk,jiang2021talk,tu2023style,otberdout2023generating}, aiming to blend explicit control with high-quality synthesis. Yet, these approaches often rely on latent space inversion \cite{zhuang2021enjoy}, which can degrade identity fidelity and blur high-frequency details \cite{tang2022facial,cheng2022inout}.

\IEEEpubidadjcol

Recent diffusion-based methods have shown strong potential for controllable face generation due to their powerful priors and high image quality \cite{park2015consensus,wei2023elite,11251121}. A popular trend is to adapt foundation models like Stable Diffusion (SD) for face editing by introducing plug-and-play components that extend its capabilities \cite{chang2023magicpose,hu2024animate}. These methods often leverage networks such as AppearanceNet \cite{hu2024animate} and CLIP \cite{radford2021learning} to extract identity features and encode editing conditions. To adapt SD to new tasks while preserving its generative ability, many approaches \cite{chang2023magicpose,han2024face,varanka2024towards} employ external control structures, such as ControlNet \cite{zhang2023adding} or adapter-based tuning strategies \cite{ye2023ip,mou2024t2i}, which modify a small subset of the model parameters. Depending on the target attribute, facial landmarks are commonly used to control expression \cite{chang2023magicpose,wei2024aniportrait}, while spherical harmonics (SH) coefficients are adopted to represent lighting conditions \cite{hou2022face,hou2021towards}.

However, despite their effectiveness, these approaches still face notable limitations. First, \textbf{the representations used for controlling facial conditions often lack robustness}: landmarks are sparse and fail to capture subtle expression changes, while low-order SH coefficients provide overly abstract lighting cues, especially in multi-attribute editing scenarios. Second, \textbf{identity preservation remains challenging}. Feature extractors like CLIP \cite{radford2021learning}, while effective for semantic understanding, tend to overlook fine-grained facial details crucial for maintaining identity fidelity. Third, due to the domain gap between the original SD training data and the face editing domain, \textbf{methods that rely on partial fine-tuning such as adapter-based \cite{ye2023ip,mou2024t2i} or ControlNet-based \cite{zhang2023adding} architectures struggle to achieve full-domain adaptation}. As a result, generated images often suffer from visual artifacts or inherit stylistic biases from the original SD model, particularly when applied to out-of-domain portraits like cartoons, sculptures, or comics.

\begin{figure*}[t] 
\centering 
\includegraphics[width=1.0\textwidth]{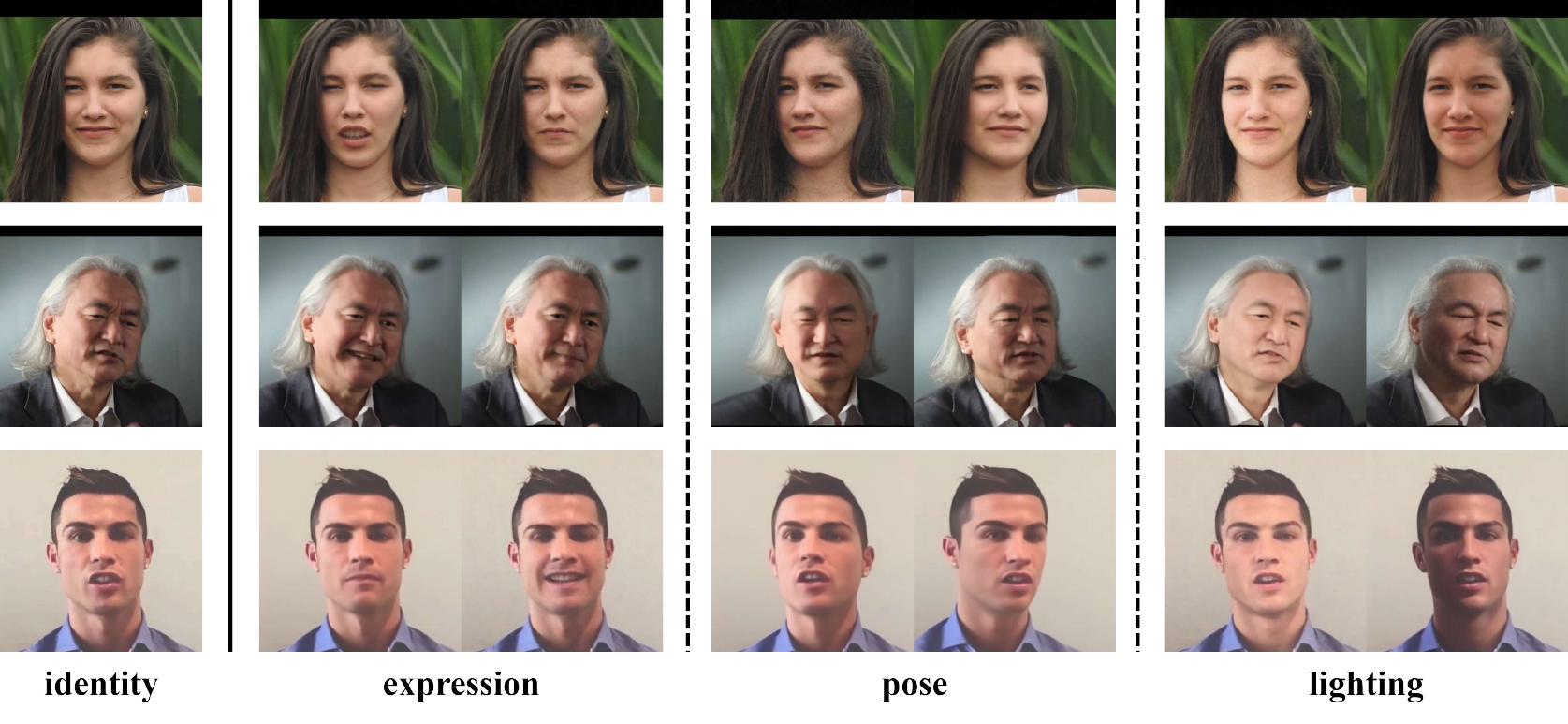}
\caption{\textbf{Consistent and controllable face editing results given identity images.} Our approach is capable of editing arbitrary identities with new facial expression, pose and lighting, generating clear and stable results while maintaining consistency with the attributes unintended to change.}
\label{fig:demo}
\end{figure*}

To address the challenges outlined above, we propose an effective framework, coined as RigFace, for photorealistically editing facial expressions, head pose, and lighting in a given portrait by combining the strengths of 3DMM and SD. Our model comprises several effective technologies: (1) Unlike adapter-based or ControlNet-style methods that only fine-tune a limited set of parameters, our approach fully fine-tunes the entire SD model, using its pre-trained weights solely for initialization. This allows the model to better adapt to our task while leveraging the powerful generative prior of SD. (2) We design a 3DMM-based Spatial Attribute Provider that automatically generates 3D renderings, masked backgrounds, and expression parameters, enabling more continuous expression representation and more accurate modeling of complex lighting variations. (3) To enhance identity preservation, we introduce the FaceFusion module, which transfers features from the Identity Encoder to the Denoising UNet through layer-wise injection into the self-attention of transformer blocks. This design significantly improves the consistency and fidelity of identity-specific details in the generated results.


Through the aforementioned improvements, our approach is capable of effectively editing freestyle portraits, as illustrated in the generated images of Fig. \ref{fig:demo} and \ref{fig:comp1}. We evaluate our method across a wide range of face images with diverse forms and styles. The results demonstrate its strong generalization ability in handling portraits and conditions that lie outside the training domain. Compared to existing baseline methods, our approach achieves superior performance both quantitatively and qualitatively, delivering high visual fidelity, accurate identity preservation, and precise condition rendering. In summary, our contributions are as follows:

\begin{itemize}
  \item We introduce RigFace, a face editing model to facilitate precise control over head pose, lighting and facial expressions for a given identity. This model proficiently enables face editing, surpassing previous GAN-based and diffusion-based methods.
  \item We propose an innovative Spatial Attribute Provider that separately generate 3D renderings, target background, and expression parameters as the disentangled conditions. This allows the model to learn the mapping from specific conditions to edited images.
  \item We propose FaceFusion to efficiently align the identity features with the target face in the self-attention of transformer blocks without redundant warping.
  \item RigFace is powerful by inheriting the knowledge from pre-trained latent diffusion models. Releasing all parameters can unleash the full potential of the model, enabling the SD model to achieve better adaptability to our task.
\end{itemize}

\vspace{-0.5em}

\section{Related Work}

\subsection{Facial Appearance Editing}
Facial Appearance Editing focuses on modifying a source face by leveraging the attribute cues extracted from a reference or query image. Due to the intrinsic disentanglement capabilities of 3D Morphable Models (3DMMs) for identity, facial expression, pose, and lighting, editing attributes such as expression, lighting, and head pose has long been a standard practice in 3D face modeling. However, directly rendering faces with 3DMM often results in unrealistic visual quality because of limited geometric detail and fixed rendering assumptions \cite{feng2021learning,deng2019accurate,hong2022headnerf,huang2015identifying}. To overcome this limitation, many approaches incorporate 3DMMs as controllable conditions into generative models to produce more photorealistic editing results. For example, in early works, the GAN-based StyleRig \cite{tewari2020stylerig} utilizes 3DMM parameters to drive face editing on a pre-trained StyleGAN \cite{karras2020analyzing}. HeadGAN \cite{doukas2021headgan} extracts and adapts 3D head representations from driving videos. However, GAN-based approaches typically depend on latent space inversion to project real images into the GAN’s semantic space \cite{zhuang2021enjoy}, which often results in identity mismatch and the degradation of fine-grained facial details. In recent years, with the rise of diffusion models demonstrating superior fidelity in image synthesis, increasing attention has been given to integrating 3DMMs with diffusion-based architectures for high-quality facial appearance editing \cite{wei2024aniportrait,han2024face,ding2023diffusionrig}. Although these methods have significantly advanced the field, key challenges outlined in Sec. \ref{sec:intro} remain insufficiently addressed: suboptimal identity preservation, limited generation fidelity, and inaccurate attribute transfer. To address these issues, we introduce a new pipeline for facial appearance editing that effectively tackles the three aforementioned challenges.


\subsection{Personalization of Pretrained Diffusion Models}

Personalization in text-to-image diffusion models typically involves adapting pre-trained backbones to specific tasks. Early approaches \cite{gal2022image,ruiz2023dreambooth,wei2024dreamvideo} achieved personalization by employing optimization or fine-tuning techniques based on text inversion \cite{gal2022image}, typically using an image collection of a specific subject. While effective for capturing subject-specific features, these methods suffer from limited generalization, as the trained model is tightly bound to the identity it was fine-tuned on. To improve efficiency, more recent methods such as T2I-Adapter \cite{mou2024t2i}  and ControlNet \cite{zhang2023adding} introduced parameter-efficient tuning by adapting SD either through updating only a subset of its parameters or by incorporating external control mechanisms. While this improves generalization by eliminating the need for subject-specific fine-tuning, it comes at the expense of visual realism. Adapter fine-tuning more generally constrains the model’s ability to adapt to new tasks that diverge significantly from the original text-to-image objective. For example, FaceAdapter \cite{han2024face} and FineFace \cite{varanka2024towards}, both adopting adapter strategies, tend to produce unrealistic lighting and facial artifacts. Similarly, ControlNet-based methods, although effective in certain tasks \cite{chen2023controlstyle,chen2025unirestore,lin2024diffbir}, often struggle to preserve fine facial details in face-specific applications and exhibit reduced generalization performance when applied to out-of-domain portraits \cite{chang2023magicpose}.

To overcome these limitations, RigFace adopts a full fine-tuning strategy on pre-trained diffusion models to enable better adaptation to the specific requirements of identity-aware face editing. Although this strategy involves higher memory usage compared to parameter-efficient approaches, it benefits from the use of pre-trained weights for faster convergence. In addition, unlike methods that encode identity information through CLIP embeddings \cite{varanka2024towards} or adapter modules \cite{mou2024t2i,han2024face}, RigFace introduces a dedicated FaceFusion module that integrates high-fidelity identity features directly into the SD architecture. Despite the full fine-tuning, training remains efficient and can be completed within 24 GPU hours on a single AMD MI250X GPU. This design enables RigFace to achieve superior identity preservation, editability, and visual realism without the typical trade-offs of parameter-efficient tuning.


\section{Method}

\subsection{Preliminaries}

\noindent\textbf{3D Morphable Face Models.} 3D Morphable Face Models (3DMMs) are parametric models that utilize a compact and entirely disentangled latent space, either handcrafted or learned from scans, to encode attributes such as head pose, facial geometry, and expressions \cite{tellamekala20233d,sun2023continuously,lee2014intra}. In this paper, we adopt FLAME \cite{li2017learning} and BFM \cite{paysan20093d}, two widely used 3DMMs that employ standard vertex-based linear blend skinning with corrective blendshapes, constructing facial meshes through pose, shape, and expression parameters. While these models offer a compact and physically meaningful representation of facial geometry, it lacks the description for appearance from face images. To address this, DECA \cite{feng2021learning} builds upon FLAME \cite{li2017learning} by incorporating Lambertian reflectance and Spherical Harmonics (SH) lighting to model facial appearance. In our work, we leverage DECA \cite{feng2021learning} to generate rough 3D representations, enabling flexible "rigging" of pose and lighting by modifying FLAME \cite{li2017learning} parameters. For facial expression rigging, we did not use FLAME \cite{li2017learning} because its 3D template ($\sim5,000$ vertices) lacks the ability to capture detailed facial variations. Instead, we opted for Deep3DRecon \cite{deng2019accurate}, which is based on the BFM \cite{paysan20093d} template ($\sim35,000$ vertices) to extract expression parameters from images. As illustrated in Fig. \ref{fig:real}, Deep3DRecon \cite{deng2019accurate} indeed performs better in expressing facial expressions than DECA \cite{feng2021learning}, and there is a noticeable realism gap between rendered images and real photos, highlighting the need for post-editing adjustments.

\begin{figure*}[htbp]
    \centering
    \includegraphics[width=0.95\textwidth]{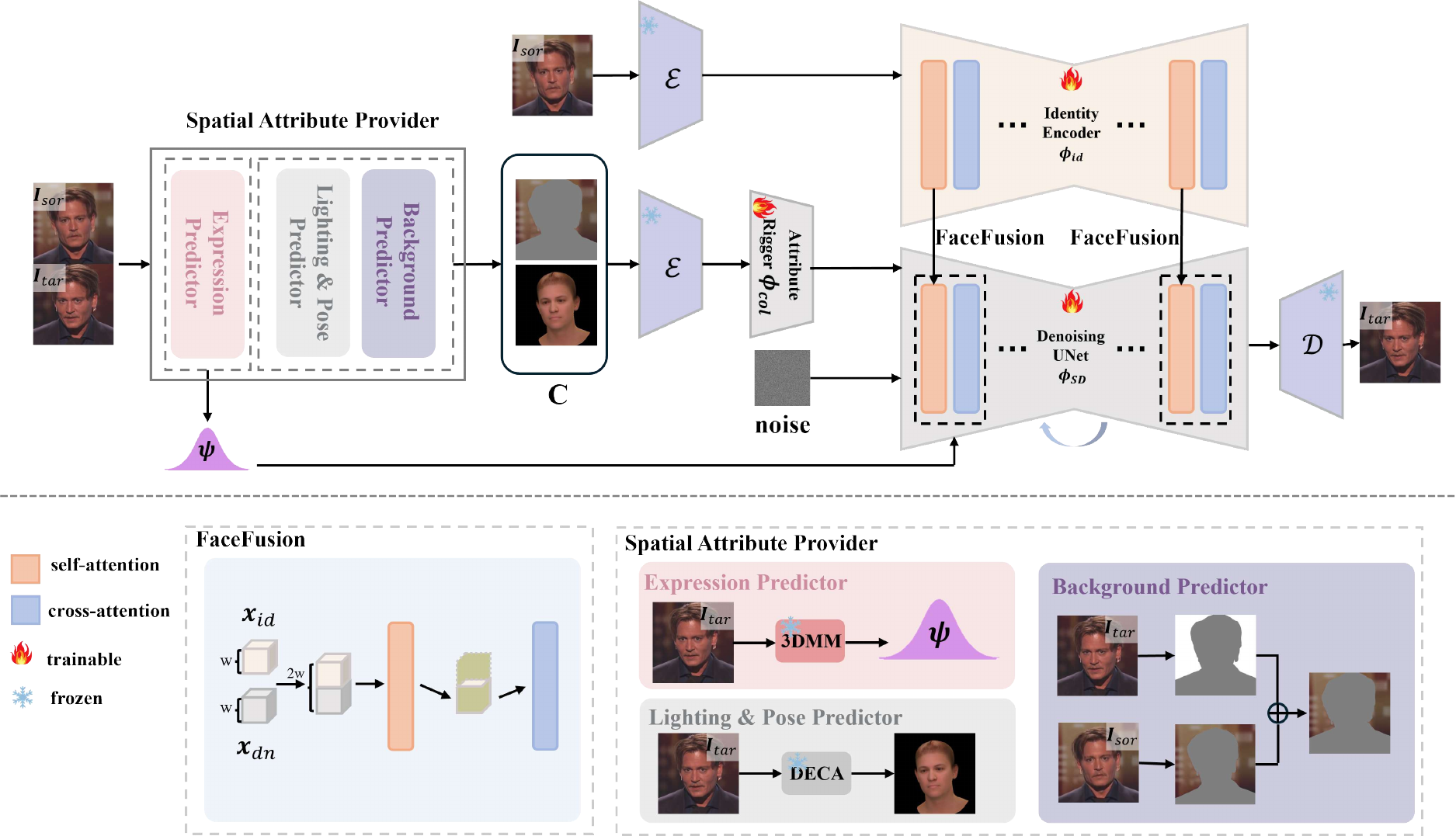}
    \caption{\textbf{Overview pipeline of RigFace.} The Spatial Attribute Provider adapts the foreground mask and predict 3D rendering as well as expression coefficients, offering decoupled and more clear guidance for controlled generation. The mask and rendering are encoded using Attribute and fused with noise, followed by the Denoising UNet conducting the denoising process for generation. The expression coefficients are directly encoded in the Denoising UNet. FaceFusion involves extracting detailed features from source image (identity) through Identity Encoder and utilized for Self-Attention. The Identity Encoder and Denoising UNet are completely fine-tuned to better adapt the prior knowledge of SD to our task.}
    \label{fig:framework}
\end{figure*}

\noindent\textbf{Stable Diffusion.} In this paper, we develop our method based on the recent T2I diffusion model, specifically Stable Diffusion (SD) \cite{rombach2022high}. SD is a latent diffusion model (LDM) that consists of an autoencoder and a UNet denoiser. The autoencoder encodes an image $\mathbf{x}_0$ into the latent space as $\mathbf{z}_0$, which can then be reconstructed. The diffusion process takes place within the latent space using a modified UNet denoiser. The optimization process is formally defined by the following equation:

\begin{equation}
\mathcal{L}=\mathbb{E}_{\mathbf{z}_t, \mathbf{C}, \epsilon, t}\left(\left\|\epsilon-\epsilon_\theta\left(\mathbf{z}_t, t, \mathbf{C}\right)\right\|_2^2\right)
\end{equation}

Here, $\textbf{z}_t$ represents the noised latent at timestep $t$, while $\mathbf{C}$ denotes the conditional text embedding generated by the pre-trained CLIP \cite{radford2021learning} text encoder. The function $\epsilon_\theta$ refers to the UNet denoiser. During the sampling process, the latent 
$\textbf{z}_t$ is progressively denoised from an initial random Gaussian noise using $\epsilon_\theta$, which is conditioned on both $\textbf{C}$ and $t$. Finally, the denoised latent is decoded into an image by the autoencoder’s decoder.

\subsection{Model Overview}

The structure of the proposed method is illustrated in Fig. \ref{fig:framework}, which aims to assemble identity clues with new pose, expression and lighting attributes provided by the spatial attribute provider. 

Since the ground truth for edited images is unavailable, we train the model using image pairs of the same identity but with varying pose, expression, and background. The model learns to transform the source image $\boldsymbol{I}_{sor}$ into the target image $\boldsymbol{I}_{tar}$ based on the injected conditions. During training, all conditions are extracted from the target image, while at inference, the background comes from the identity image. The Spatial Attribute Provider selectively adjusts the desired editing conditions, keeping others consistent with the identity image.

\subsection{Spatial Attribute Provider}

To provide a precise guidance for subsequent controllable generation, we design a Spatial Attribute Provider (SAP) to predict varying background area, rendering, and expression coefficients which represent decoupled pose, lighting and expression conditions.

\noindent \textbf{Expression Predictor.} We utilize Deep3DRecon \cite{deng2019accurate} to extract expression coefficients $\boldsymbol{\psi}$ of the target faces. Although this model provides other coefficients related to pose and lighting, we find that it doesn't work to change the pose and lighting if these coefficients are applied, we hypothesize this is because that pose and lighting require more intuitive and low-level guidance while coefficients are over high-level to make model understand, which we will validate in Sec. \ref{sec:abl}.

\noindent \textbf{Lighting \& Pose Predictor.} As mentioned above, to provide intuitive guidance about lighting and pose, we use the rendering image from DECA \cite{feng2021learning} which is pixel-aligned with the target face. Specifically, DECA \cite{feng2021learning} produces the physical coefficients of lighting (spherical harmonics) and angles from the target image. We then use the Lambertian reflectance to render these physical coefficients into the Lambertian rendering.

\begin{figure}[htbp]
    \centering
    \includegraphics[width=0.45\textwidth]{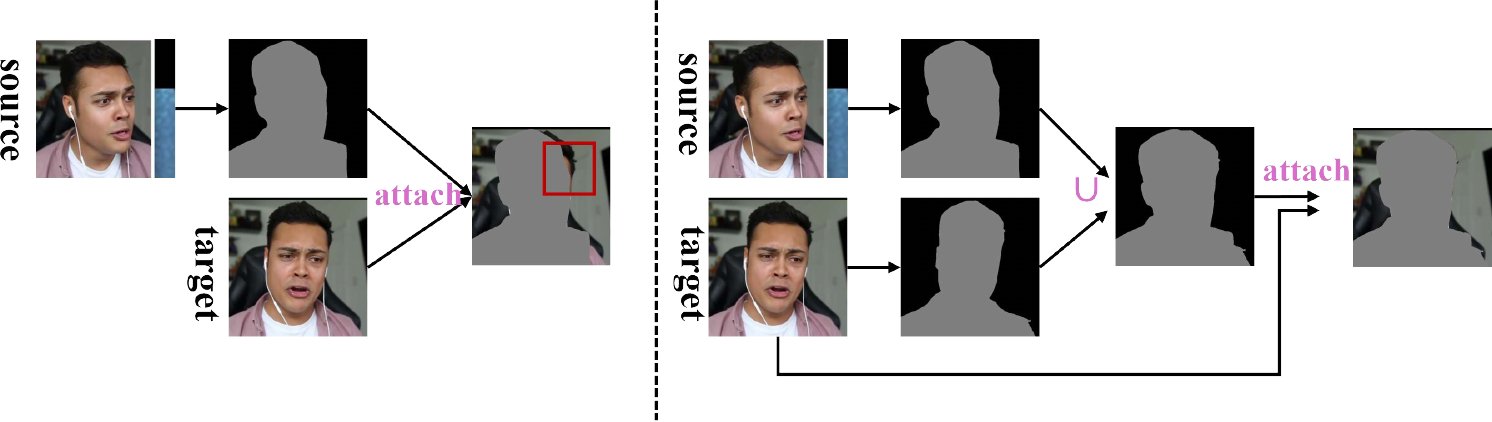}
    \caption{\textbf{Illustration of how the background is parsed based on the source and target images in the training data.} The left panel shows the result of attaching only the masked-out facial region from the source image onto the background of the target image. The right panel demonstrates attaching the source facial region using a combination of the source and target masks to better align with the target background. The \textcolor{red}{red box} highlights regions of target identity leakage, where residual facial information from the target image remains visible.}
    \label{fig:add}
\end{figure}

\noindent \textbf{Background Predictor.} To address the issue of background changes from the source image to the target image, the model needs to be explicitly informed of current background of the target. Incorporating the background as a constraint greatly reduces the complexity of the model's learning process, shifting it from generating entirely new images to performing conditional inpainting. As a result, the model becomes better adapted to maintaining background consistency and producing content that integrates smoothly with it.

Specifically, we use a pre-trained face parsing model \cite{yu2021bisenet} to obtain the head region. Due to the limitations of in-the-wild datasets, it is generally difficult to obtain training samples where only a single attribute (e.g., expression) changes while head pose and lighting remain fixed. During the training phase, most of the source–target image pairs we can construct involve simultaneous changes in multiple factors, including pose, expression, and lighting. As a result, we assume that all three conditions are edited jointly during training. In such settings, directly applying the head mask of the target image to the identity image is often impractical, because the target image typically contains residual identity-specific features that differ from the source. As illustrated in Fig. \ref{fig:add}, considering only the source pose may leave behind unwanted identity information from the target image, leading to visual artifacts (highlighted in the red box of the left panel). To address this, we extract the head regions from both the source and target images and combine them into a new image, allowing better alignment with the target background (right panel). The result image is then used to mask the foreground of the target image, which is one of the conditions sent into Attribute Rigger in Fig. \ref{fig:framework}. At inference time, for expression or lighting editing (without pose adjustment), we can safely use the source mask only, as there is no spatial misalignment. In contrast, for pose editing, we compute the union of the source and target pose masks and use this combined region for face compositing. Since our model is trained with such mask unions, it learns to inpaint the necessary background around the new head region, enabling it to handle moderate spatial shifts effectively.

\subsection{FaceFusion}

To preserve the facial details of the identity image, we need to integrate the face information from the source image into the diffusion U-Net. Although many SD-based methods use CLIP \cite{radford2021learning} as the reference image encoder \cite{han2024face,varanka2024towards}, we find that it fails to preserve many detailed identity features, as validated in Sec. \ref{sec:abl}. Previous works \cite{hu2024animate,liu2024towards} have demonstrated that self-attention effectively preserves reference image content through information fusion. Inspired by this, we design FaceFusion to efficiently learn the detail features of the face characteristics from the Identity Encoder. The left side of Fig. \ref{fig:framework} shows the architecture of the Identity Encoder, which is essentially identical to the Denoising UNet of SD. It only inherits the weights in the SD model for initialization, and during the training all parameters are released. Given the source image $\mathbf{I}_{sor}$, a frozen VAE encoder $\mathcal{E}$ compresses the image into a low dimension feature representation and then sent it into the Indentity Encoder. The Denoising UNet is also inherited from the SD model with all parameters released. As demonstrated in the FaceFusion blocks of Fig. \ref{fig:framework}, in each transformer blocks of these two models, suppose feature $\mathbf{x}_{id}\in \mathbb{R}^{h \times w \times c}$ from the Identity Encoder and $\mathbf{x}_{dn} \in \mathbb{R}^{h \times w \times c}$ from the Denoising UNet, $\mathbf{x}_{id}$ is first concatenated with $\mathbf{x}_{dn}$ along the $w$ dimension. Then the self-attention layer takes in it and only half feature of the output along the $w$ dimension is kept and sent into the cross-attention layer. This module plays a crucial role in preserving high-frequency identity details in the generated images, ensuring faithful identity consistency throughout the editing process.



\subsection{Training Strategy}

We design the Attribute Rigger to be lightweight by utilizing a convolution layer with 8 channels, $4\times4$ kernels and $2\times2$ strides. To be specific, the VAE encoder $\mathcal{E}$ in SD compresses the spatial conditions into latent features and then the Attribute Rigger takes in the features and integrates them into the Denoising UNet $\boldsymbol\phi_{SD}$. The expression coefficients $\boldsymbol{\psi}$ are mapped by a linear layer to have the same dimension with the time embedding of the Denoising UNet and then added to the time embedding. As a result, given all the conditions and identity features, the whole framework is optimized by minmizing the following:

\begin{equation}
\mathcal{L}=\mathbb{E}_{\mathbf{z}_t, \mathbf{g}, \mathbf{y}, \boldsymbol{\psi}, \epsilon, t}[\|\epsilon-\boldsymbol\phi_{SD}(\mathbf{z}_t, t, \boldsymbol\phi_{id}(\mathbf{g}), \boldsymbol{\psi}, \boldsymbol\phi_{col}(\mathbf{y}))\|_2^2],
\end{equation}

\noindent where $\mathbf{z}_t$ is the noised latent at timestep $t$, $\mathbf{g}=\mathcal{E}(\boldsymbol{I}_{sor})$ represents the latent feature sent to the Identity Encoder, $\mathbf{y}=\mathcal{E}(\boldsymbol{C})$ represents the latent feature sent to the Attribute Rigger, $\boldsymbol{C}$ denotes conditional images, $\boldsymbol\phi_{id}$ represents the Identity Encoder, $\boldsymbol\phi_{col}$ represents the Attribute Rigger and $\epsilon$ is the predicted noise, respectively. During both training and inference, only the Denoising UNet is involved in diffusion-related operations (i.e., noise addition and denoising). In contrast, the Identity Encoder solely processes the reference identity image to extract multi-level self-attention features, which are fused into the Denoising UNet via the FaceFusion module. Therefore, the Identity Encoder does not introduce significant computational overhead at both stages.

\begin{figure*}[htbp]
    \centering
    \includegraphics[width=1.\textwidth]{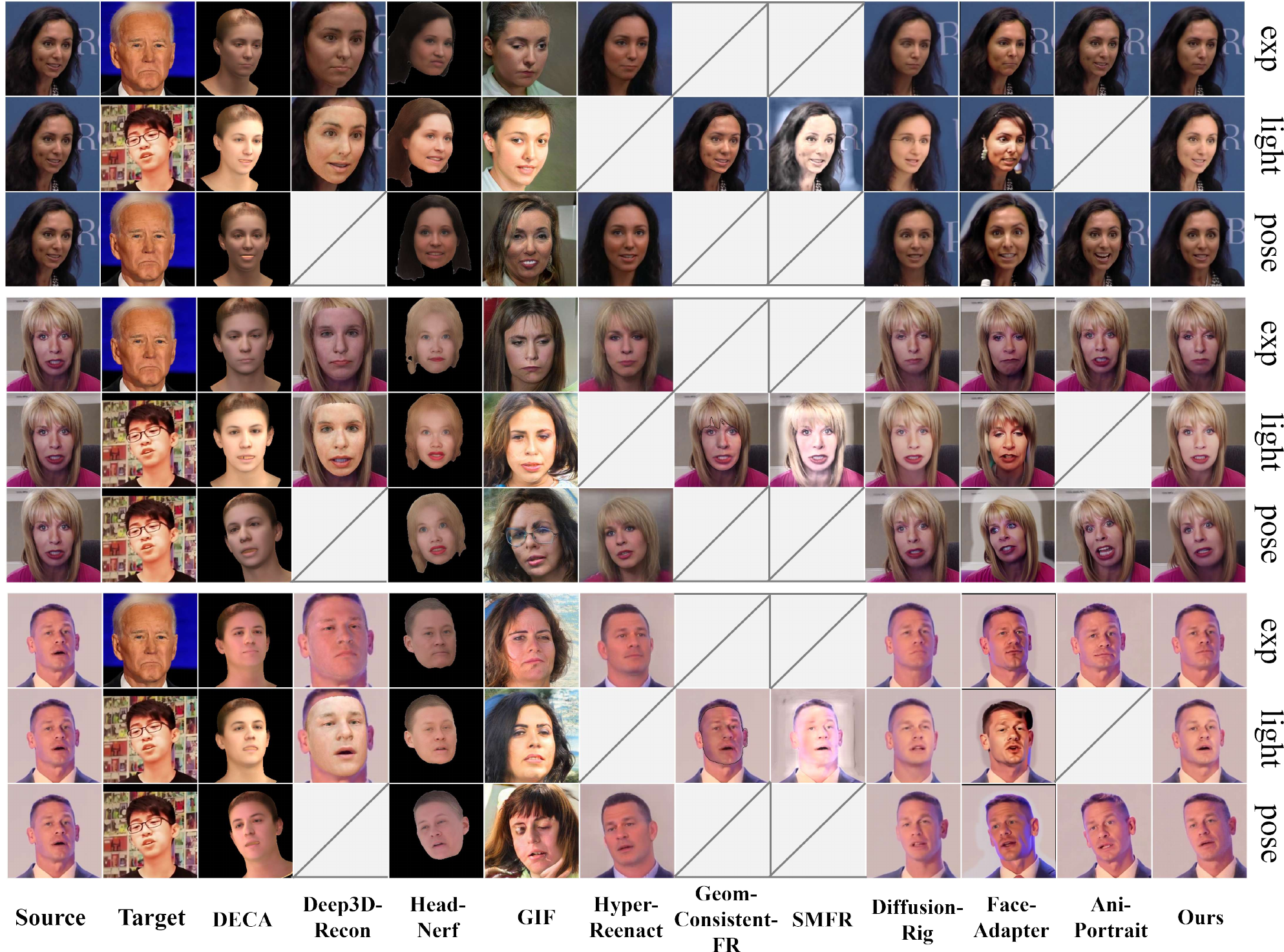}
    \caption{\textbf{Qualitative comparison between RigFace and other baselines on real face images.} The slashed cells indicate unsupported editing scenarios. Zoom in to view image details.}
    \label{fig:real}
\end{figure*}

\section{Experiments}
\subsection{Experimental Setup}
\textbf{Dataset.} We process the Aff-Wild \cite{kollias2019deep} video dataset to meet our training requirements. Aff-Wild is a large-scale in-the-wild dataset consisting of videos with a variety of subjects in different emotional states, head poses and illumination conditions. For the training set, we first decompose each video into frame sequences. From each sequence, we randomly sample a pair of frames that are temporally distant to ensure significant variation. One frame is designated as the source and the other as the target. We ensure that each video sample is used multiple times during sampling. In total, this results in approximately 30K image pairs extracted from the entire dataset for training. During the evaluation, we sampled 20 identities from the remaining Aff-Wild videos, and additionally selected 20 identities from other datasets, including CelebV-HQ \cite{zhu2022celebv}, AFEW \cite{dhall2011acted}, and AffectNet \cite{mollahosseini2017affectnet}. For each identity, we applied three types of editing, i.e., expression, pose, and lighting with two test cases per condition, resulting in a total of 360 generated images. The videos used in training and test are different to ensure reliability. For preprocessing of the sampled images, we follow FOMM \cite{siarohin2019first} to crop the face, preserve partial backgrounds, and resize them into $512\times512$ to fit in the resolution of Stable-Diffusion.

\noindent\textbf{Evaluation Criteria.} We use identity feature $\ell_2$ distance computed by a pre-trained model \footnote{https://github.com/ageitgey/face\_recognition} to evaluate identity preservation. We use SSIM \cite{wang2004image}, LPIPS \cite{zhang2018unreasonable} and FID \cite{heusel2017gans} to evaluate the perceptual quality of the edited results and background preservation. Note that for fair evaluation, these scores are computed only on the lighting and expression editing results after masking out the facial region when comparing with the ground truth. This is because pose changes affect the background content, while lighting and expression changes primarily alter facial regions, so we exclude these interfering factors. To assess the accuracy of head pose edit, we calculate the Average Pose Distance (APD), which is the mean $\ell_1$ distance between the pose parameters extracted from the animated and driving images using the SMIRK \cite{retsinas20243d} model. Additionally, we compute the Pose Root Mean Square Error (P-RMSE) using the pose coefficients predicted by Deep3DFace \cite{deng2019accurate}. For evaluating facial expression edit, we use the Average Expression Distance (AED), defined as the mean $\ell_1$ distance between the expression features extracted by the pre-trained EMOCA \cite{danvevcek2022emoca} model. We also report the Expression Root Mean Square Error (E-RMSE) based on expression coefficients predicted by Deep3DFace \cite{deng2019accurate}. For lighting edit, we use the Average Lighting Distance (ALD) computed as the mean $\ell_1$ distance between the Spherical Harmonics (SH) coefficients estimated from DECA \cite{feng2021learning}, and the Lighting Root Mean Square Error (L-RMSE) based on SH coefficients estimated by Deep3DFace \cite{deng2019accurate}. These metrics have been widely used in prior studies on face reenactment \cite{bounareli2023hyperreenact,han2024face}, facial appearance editing \cite{ding2023diffusionrig,ghosh2020gif}, and face animation \cite{guo2024liveportrait,wei2024aniportrait}, and are considered well-suited for assessing the accuracy of the respective editing effects. For methods that do not support certain editing conditions, we only include the results corresponding to the conditions they support in the evaluation.

\begin{table*}[htbp]
 \small
  \centering
  \caption{\textbf{Quantitative comparison for face editing of facial expression, lighting, and pose on real face images.} The best and second best results are reported in \textbf{bold} and \underline{underlined}, respectively. N/A indicates that the method does not support editing under the corresponding condition. Deep3DRe. denotes Deep3DRecon \cite{deng2019accurate} , and Geom. denotes GeomConsistentFR \cite{hou2022face}.}

\resizebox{\textwidth}{!}{
  \begin{tabular}{lccccccccccc}
    \toprule
    \multirow{2}{*}{Method~~} &\multicolumn{1}{c}{\textbf{ID}} &\multicolumn{3}{c}{\textbf{Image}} &\multicolumn{2}{c}{\textbf{Pose}} &\multicolumn{2}{c}{\textbf{Exp}} &\multicolumn{2}{c}{\textbf{Light}}\\

    \cmidrule(r){2-2} \cmidrule(r){3-5} \cmidrule(r){6-7} \cmidrule(r){8-9} \cmidrule(r){10-11} 
    & $\ell_2$\(\downarrow\)~~ & SSIM\(\uparrow\)~~ & LPIPS\(\downarrow\)~~  &FID\(\downarrow\)~~ & APD\(\downarrow\)~~ &P-RMSE\(\downarrow\)~~ & AED\(\downarrow\)~~ & E-RMSE\(\downarrow\)~~ & ALD\(\downarrow\) & L-RMSE\(\downarrow\)\\
    \midrule
    DECA \cite{feng2021learning}       &0.651~~ &0.558~~ & 0.622~~ &123.87~~ &\bf{0.026}~~ &\bf{0.063}~~  &0.068~~  &0.204 &\bf{0.045} &\bf{0.136}\\
    Deep3DRe. \cite{deng2019accurate} &0.588~~ &0.773~~ &0.404~~ &96.63~~ &N/A~~  &N/A ~~ &\bf{0.059}~~  &\bf{0.184}  &0.063 &0.176  \\
    HeadNerf \cite{hong2022headnerf}  &0.690~~ &0.717~~ & 0.525~~ &105.09~~ &0.042~~ &0.104~~ &0.073~~ &0.212  & 0.088 & 0.211\\
    
    GIF \cite{ghosh2020gif}     &0.624~~ &0.650~~ & 0.527~~ &86.41~~ &0.033~~ &0.086~~  &0.066~~  &0.196  & 0.050  & 0.147\\
    HyperReenact \cite{bounareli2023hyperreenact}  &0.584~~ &0.689~~ &0.308~~ &72.77~~ &0.046~~ &0.110~~ &0.074~~  &0.225 &N/A  &N/A \\
    
    Geom. \cite{hou2022face}  &0.369~~ &0.795~~ &0.112~~ &70.29~~ &N/A ~~ &N/A ~~ &N/A~~  &N/A & 0.060  &0.182\\
    SMFR \cite{hou2021towards} &0.362~~ &0.782~~ &0.114~~ &74.06~~ &N/A ~~ & N/A ~~ &N/A~~ &N/A &0.094   &0.233\\

    DiffusionRig \cite{ding2023diffusionrig} &0.455~~ &0.724~~ &0.133~~  &90.52~~ &0.028~~ &0.077~~ &0.071~~ &0.216 & 0.057 &0.169\\
    FaceAdapter  \cite{han2024face}   &0.421~~ &0.785~~ &0.125~~ &88.32~~ &0.030~~ &0.085~~ &0.063~~  &0.194 & 0.082 & 0.207\\
    AniPortrait  \cite{wei2024aniportrait}  &\underline{0.361}~~ &\underline{0.802}~~ &\underline{0.090}~~  &\bf{49.55}~~ &0.035~~ &0.088~~ &0.066~~ &0.200 &N/A  &N/A \\
    \midrule
    \textbf{Ours} ~~  &{\bf 0.356}~~ &{\bf 0.850}~~ &{\bf 0.063}~~ &\underline{60.27}~~ &\bf{0.026}~~ &\underline{0.069}~~ &\underline{0.061}~~ &\underline{0.188} & \underline{0.053}  & \underline{0.142}\\
    \bottomrule 
  \end{tabular}}
  \label{tab:qua_in}
\end{table*}

\begin{table*}[htbp]
 \small
  \centering
  \caption{\textbf{Quantitative comparison for face editing of facial expression, lighting, and pose across multiple face image styles.} The best and second best results are reported in \textbf{bold} and \underline{underlined}, respectively. N/A indicates that the method does not support editing under the corresponding condition. Deep3DRe. denotes Deep3DRecon \cite{deng2019accurate} , and Geom. denotes GeomConsistentFR \cite{hou2022face}.}
\resizebox{\textwidth}{!}{
  \begin{tabular}{lccccccccccc}
    \toprule
    \multirow{2}{*}{Method~~} &\multicolumn{1}{c}{\textbf{ID}} &\multicolumn{3}{c}{\textbf{Image}} &\multicolumn{2}{c}{\textbf{Pose}} &\multicolumn{2}{c}{\textbf{Exp}} &\multicolumn{2}{c}{\textbf{Light}}\\

    \cmidrule(r){2-2} \cmidrule(r){3-5} \cmidrule(r){6-7} \cmidrule(r){8-9} \cmidrule(r){10-11} 
    & $\ell_2$ \(\downarrow\)~~ & SSIM\(\uparrow\)~~ & LPIPS\(\downarrow\)~~  &FID\(\downarrow\)~~ & APD\(\downarrow\)~~ &P-RMSE\(\downarrow\)~~ & AED\(\downarrow\)~~ & E-RMSE\(\downarrow\)~~ & ALD\(\downarrow\) & L-RMSE\(\downarrow\)\\
    \midrule
    DECA \cite{feng2021learning}       &0.634~~ &0.549~~ & 0.628~~ &139.03~~ &\bf{0.024}~~ &\bf{0.065}~~  &0.072~~  &0.192 & \bf{0.052} & \bf{0.147}\\
    Deep3DRe. \cite{deng2019accurate} &0.613~~ &0.760~~ &0.412~~ &105.54~~ &N/A ~~ &N/A~~ &\bf{0.054}~~ &\bf{0.170}  &0.068  &0.169 \\
    HeadNerf \cite{hong2022headnerf}  &0.650~~ &0.704~~ & 0.531~~ &112.59~~ &0.055~~ &0.117~~ &0.090~~  &0.234 & 0.102 &0.249\\
    
    GIF \cite{ghosh2020gif}     &0.629~~ &0.652~~ & 0.535~~ &99.29~~ &0.036~~ &0.097~~  &\underline{0.059}~~  &\underline{0.173} & 0.058 & 0.157\\
    HyperReenact \cite{bounareli2023hyperreenact}  &0.611~~ &0.685~~ &0.323~~ &84.71~~ &0.051~~ &0.129~~ &0.067~~  &0.181 &N/A  &N/A \\
    
    Geom. \cite{hou2022face}  &0.388~~ &0.773~~ &0.123~~ &77.87~~ &N/A~~ &N/A~~ &N/A ~~ &N/A & 0.066 & 0.165\\
    SMFR \cite{hou2021towards} &\underline{0.377}~~ &0.761~~ &0.118~~ &77.25~~ &N/A~~ &N/A~~ &N/A ~~ &N/A & 0.097 &0.241\\

    DiffusionRig \cite{ding2023diffusionrig} &0.438~~ &0.709~~ &0.139~~  &102.09~~ &0.040~~ &0.084~~ &0.086~~ &0.224 & 0.064 & 0.160\\
    FaceAdapter  \cite{han2024face}   &0.463 ~~ &0.765~~ &0.133~~ &96.16~~ &0.037~~ &0.105~~ &0.075~~  &0.198 & 0.091  & 0.214\\
    AniPortrait \cite{wei2024aniportrait}  &0.387 ~~ &\underline{0.792} ~~ &\underline{0.102}~~  &\bf{66.33}~~ &0.046~~ &0.096~~ &0.076~~ &0.201 &N/A  &N/A \\
    \midrule
    \textbf{Ours} ~~  &{\bf 0.368}~~ &{\bf 0.839}~~ &{\bf 0.077}~~ &\underline{73.72} ~~ &\underline{0.034}~~ &\underline{0.082}~~ &0.064~~  &0.176 & \underline{0.057}  & \underline{0.152}\\
    \bottomrule
  \end{tabular}}
  \label{tab:qua_out}
\end{table*}

\begin{table*}[htbp]
 \small
  \centering
  \caption{\textbf{Quantitative evaluation of generalization ability of RigFace.} $\text{RigFace}^{\dagger}$ denotes the pipeline is directly evaluated on the small dataset after being trained on AffWild \cite{kollias2019deep}, and $\text{RigFace}^{\ddagger}$ represents the pipeline is further fine-tuned on the small dataset with diverse face styles.}
  
\resizebox{\textwidth}{!}{
  \begin{tabular}{lccccccccccc}
    \toprule
    \multirow{2}{*}{Method~~} &\multicolumn{1}{c}{\textbf{ID}} &\multicolumn{3}{c}{\textbf{Image}} &\multicolumn{2}{c}{\textbf{Pose}} &\multicolumn{2}{c}{\textbf{Exp}} &\multicolumn{2}{c}{\textbf{Light}}\\

    \cmidrule(r){2-2} \cmidrule(r){3-5} \cmidrule(r){6-7} \cmidrule(r){8-9} \cmidrule(r){10-11} 
    & $\ell_2$\(\downarrow\)~~ & SSIM\(\uparrow\)~~ & LPIPS\(\downarrow\)~~  &FID\(\downarrow\)~~ & APD\(\downarrow\)~~ &P-RMSE\(\downarrow\)~~ & AED\(\downarrow\)~~ & E-RMSE\(\downarrow\)~~ & ALD\(\downarrow\) & L-RMSE\(\downarrow\)\\
    \midrule
    $\text{RigFace}^{\dagger}$       &0.379~~ &0.818~~ & 0.080~~ &89.41~~ &\bf{0.032}~~ &\bf{0.074}~~  &0.066~~  &0.180 & 0.050 & 0.144\\
    $\text{RigFace}^{\ddagger}$    &{\bf 0.368}~~ &{\bf 0.839}~~ &{\bf 0.077}~~ &\bf{73.72}~~ &0.034~~ &0.082~~ &\bf{0.059}~~  &\bf{0.173} & \bf{0.057}  & \bf{0.152}\\
    \bottomrule
  \end{tabular}}
  \label{tab:out}
\end{table*}

\noindent\textbf{Implementation Details.} {We use the Adam optimizer \cite{kingma2014adam}. The UniPCMultistepScheduler \cite{zhao2023unipc} is employed as scheduler, which by default uses 1000 diffusion steps for training and 50 steps during inference. We train the model for 100,000 steps using 2×AMD MI250x GPUs, with a constant learning rate of 1e-5 and a batch size of 4 per GPU. The total training time is approximately 12 hours. Inference is performed on a single GPU, and the average inference time per image is around 5 seconds. Both the Identity Encoder and the Denoising UNet inherit from the Stable Diffusion 1-5 base model \footnote{https://huggingface.co/stable-diffusion-v1-5/stable-diffusion-v1-5}.

\subsection{Experimental Results}
\noindent \textbf{Comparison with SoTA methods.} We compare our method with state-of-the-art approaches both quantitatively and qualitatively on the test set. We selected representative methods from four categories: (1) 3D rendering-based methods, including DECA \cite{feng2021learning}, HeadNerf \cite{hong2022headnerf}, and Deep3DRecon \cite{deng2019accurate}; (2) GAN-based face editing methods, including GIF \cite{ghosh2020gif} and HyperReenact \cite{bounareli2023hyperreenact}; (3) face relighting methods, i.e., GeomConsistentFR \cite{hou2022face} and SMFR \cite{hou2021towards}; and (4) diffusion-based methods, including DiffusionRig \cite{ding2023diffusionrig}, FaceAdapter \cite{han2024face}, and AniPortrait \cite{wei2024aniportrait}. Among them, DiffusionRig \cite{ding2023diffusionrig} is a self-designed and fully trained diffusion model. FaceAdapter \cite{han2024face} extends IP-Adapter \cite{ye2023ip} by freezing part of the attention layers in SD. AniPortrait \cite{wei2024aniportrait} also fine-tunes the entire SD model but adopts a different encoder for identity representation and utilizes distinct expression and pose conditioning strategies. For all these methods, we use the official implementations and publicly released pre-trained weights for evaluation. Additionally, to demonstrate the high perceptual fidelity of our generated results, we conducted a user study. The details and results of this study are provided in the \hyperlink{user}{\textit{User Study}} section.

\begin{table*}[htbp]
 \small
  \centering
  \caption{\textbf{Ablation study on disentangled and pixel-aligned image condition as well as the design of Identity Encoder.} The best and second best results are reported in \textbf{bold} and \underline{underlined}, respectively.}

\resizebox{\textwidth}{!}{
  \begin{tabular}{lccccccccccc}
    \toprule
    \multirow{2}{*}{Method~~} &\multicolumn{1}{c}{\textbf{ID}} &\multicolumn{3}{c}{\textbf{Image}} &\multicolumn{2}{c}{\textbf{Pose}} &\multicolumn{2}{c}{\textbf{Exp}} &\multicolumn{2}{c}{\textbf{Light}}\\

    \cmidrule(r){2-2} \cmidrule(r){3-5} \cmidrule(r){6-7} \cmidrule(r){8-9} \cmidrule(r){10-11} 
    & $\ell_2$\(\downarrow\)~~ & SSIM\(\uparrow\)~~ & LPIPS\(\downarrow\)~~  &FID\(\downarrow\)~~ & APD\(\downarrow\)~~ &P-RMSE\(\downarrow\)~~ & AED\(\downarrow\)~~ & E-RMSE\(\downarrow\)~~ & ALD\(\downarrow\) & L-RMSE\(\downarrow\)\\
    \midrule
    No-disent.       &0.364~~ &0.729~~ & 0.081~~ &104.11~~ &0.061~~ &0.330~~  &0.109~~  &0.372 &0.139 &0.363\\
    Coef.-sep.       &0.366~~ &0.753~~ &0.070~~ &90.24~~ &0.035~~  &0.082~~ &\underline{0.071}~~  &\underline{0.218}  &0.125 &0.250  \\
    \midrule
    Control.         &0.604~~ &0.554~~ & 0.330~~ &127.85~~ &0.144~~ &0.338~~ &0.126~~ &0.345  & 0.267 & 0.492\\
    
    CLIP-ID          &0.457~~ &0.681~~ & 0.083~~ &97.93~~ &\underline{0.030}~~ &\underline{0.077}~~  &0.084~~  &0.260  & \underline{0.066}  &\underline{0.151}\\
    CLIP-light       &0.358~~ &0.745~~ &0.074~~ &87.20~~ &0.056~~ &0.208~~ &0.082~~  &0.255 &0.128  &0.260 \\
    
    Conv-ID         &\underline{0.361}~~ &\underline{0.771}~~ &\underline{0.069}~~ &78.38~~ &0.044~~ &0.103~~ &0.079~~  &0.246 & 0.085  &0.194\\
    \midrule
    \textbf{Ours} ~~  &{\bf 0.356}~~ &{\bf 0.850}~~ &{\bf 0.063}~~ &\bf{60.27}~~ &\bf{0.026}~~ &\bf{0.069}~~ &\bf{0.063}~~  &\bf{0.194} & \bf{0.053}  & \bf{0.142}\\
    \bottomrule
  \end{tabular}}
  \label{tab:abl1}
\end{table*}

\begin{figure*}[htbp]
    \centering
    \includegraphics[width=0.98\textwidth]{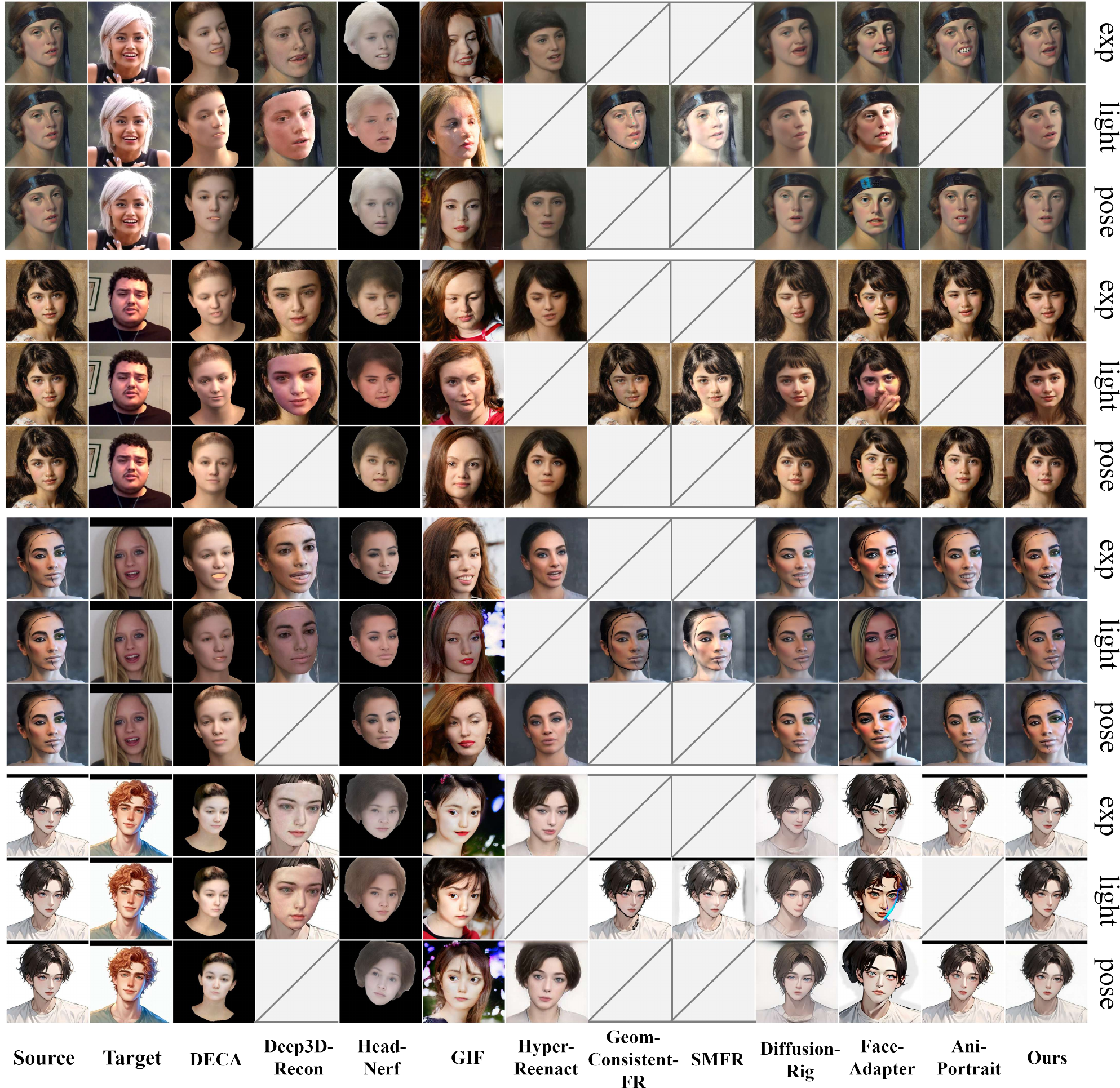}
    \caption{\textbf{Qualitative comparison between RigFace and other baselines across multiple styles.} The slashed cells indicate unsupported editing scenarios. From top to bottom, the styles are oil painting, cyberpunk-styled, cartoon and comic. Zoom in to view image details.}
    \label{fig:comp1}
\end{figure*}

\begin{figure}[H]
    \centering
    \includegraphics[width=0.48\textwidth]{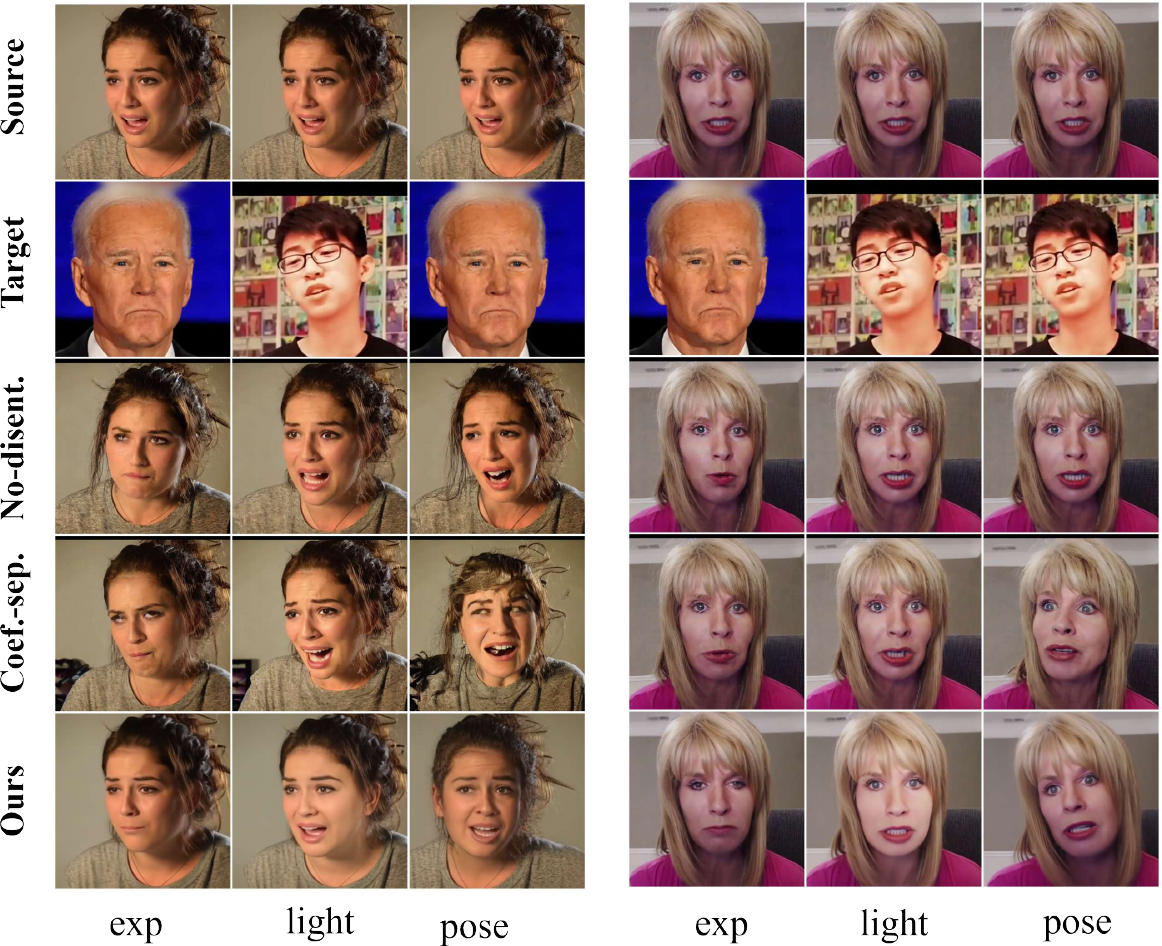}
    \caption{{\textbf{Ablation study for disentangled and pixel-aligned image condition.} \textit{No-disent.} uses only 3DMM parameters as input, jointly embedding expression, pose, and lighting coefficients without image-based cues. \textit{Coef.-sep.} also uses 3DMM parameters, but embeds expression, pose, and lighting separately to encourage disentanglement. Better viewed when zoomed in.}}
    \label{fig:abl}
\end{figure}

\begin{figure}[htbp]
    \centering
    \includegraphics[width=0.48\textwidth]{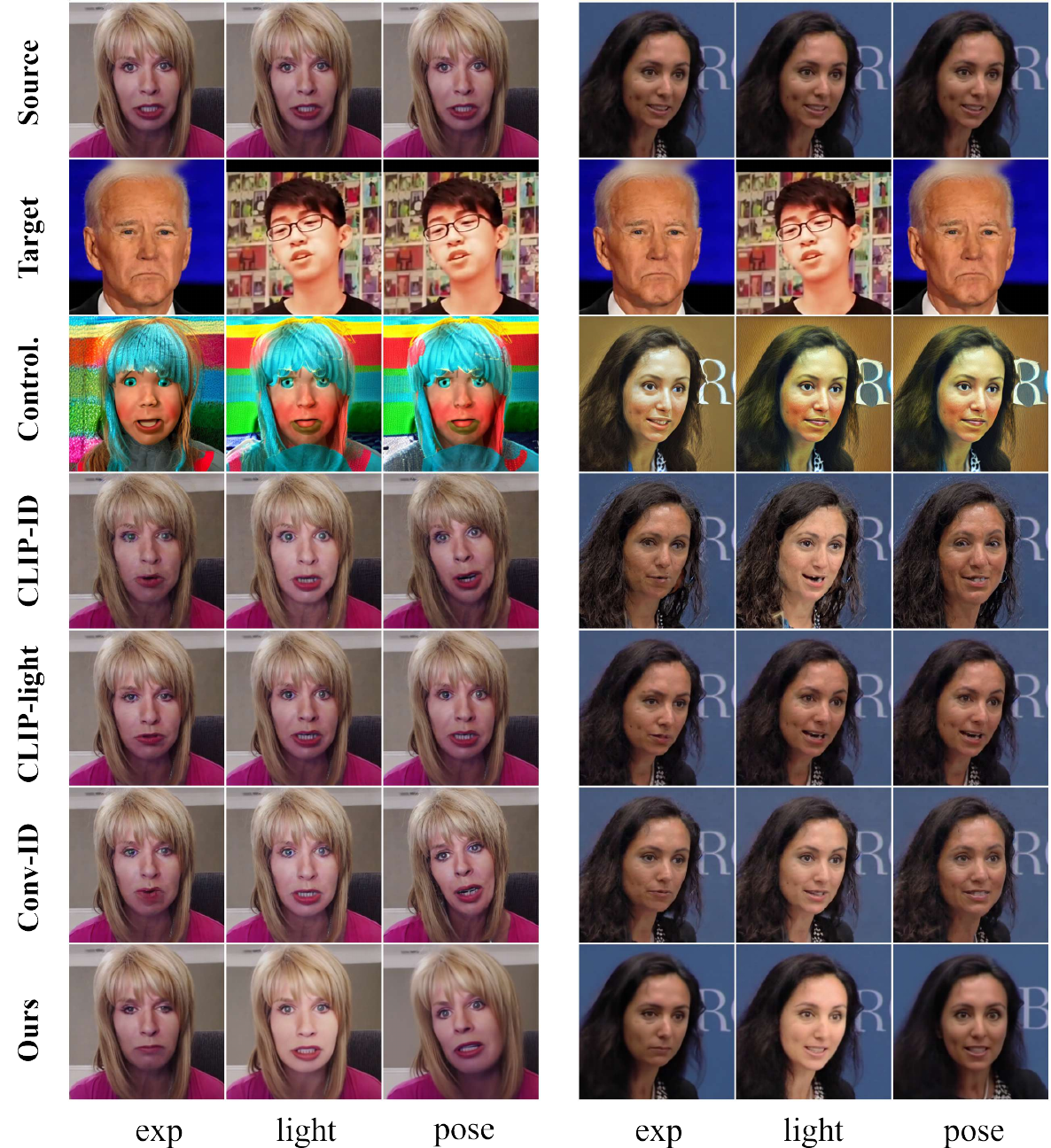}
    \caption{\textbf{Ablation study for the architecture of Identity Encoder.} \textit{Control.} uses ControlNet to embed identity. \textit{CLIP-ID} uses a CLIP image encoder for identity features. \textit{CLIP-light} applies CLIP to encode lighting conditions. \textit{Conv-ID} embeds identity via a single convolutional layer.}
    \label{fig:abl2}
\end{figure}

Fig. \ref{fig:real} presents visual comparisons between our method and other face editing approaches on several identities. We observe that, compared to other methods, RigFace consistently achieves superior editing results across a variety of identities and backgrounds. Compared to direct 3D rendering approaches such as DECA \cite{feng2021learning}, HeadNerf \cite{hong2022headnerf}, and Deep3DRecon \cite{deng2019accurate}, our method excels at synthesizing background content especially when the head pose changes, and it also produces natural-looking facial edits. Compared to GAN-based methods such as GIF \cite{ghosh2020gif} and HyperReenact \cite{bounareli2023hyperreenact}, our method achieves much better identity preservation. Compared to face relighting methods including GeomConsistentFR \cite{hou2022face} and SMFR \cite{hou2021towards}, our approach also demonstrates high accuracy in editing lightings, further validating the versatility and robustness of our method. Our method also outperforms other diffusion-based approaches in generating more realistic faces while maintaining high identity fidelity. Specifically, DiffusionRig \cite{ding2023diffusionrig} requires subject-specific album fine-tuning, leading to unstable generation quality across subjects and vulnerability to the quality of the album. For example, noticeable difference in ID preservation can be observed between the first and second subjects of Fig. \ref{fig:real}. We can also observe that adapter-based fine-tuning methods such as FaceAdapter \cite{han2024face} result in less realistic lighting and facial details, particularly in certain facial regions, due to limited model adaptation capacity. In addition, our method leverages disentangled expression coefficients as control signals, which provide a more accurate and expressive representation of facial movements compared to the sparse keypoints used in AniPortrait \cite{wei2024aniportrait}, resulting in higher expression fidelity.

Tab. \ref{tab:qua_in} presents the quantitative evaluation results of different methods. Our method shows a clear improvement in preserving identity appearance and achieving high perceptual quality,  as depicted by the large increase in LPIPS \cite{zhang2018unreasonable}, FID \cite{heusel2017gans}, and decrease in SSIM \cite{wang2004image} and identity distance scores. DECA \cite{feng2021learning} 3D rendering performs on par with ours in terms of pose and lighting metrics, and Deep3DRecon \cite{deng2019accurate} also shows comparable error in expression editing than ours, which further demonstrates the reliability of using the pose and lighting condition from DECA \cite{feng2021learning} and using the expression condition from Deep3DRecon \cite{deng2019accurate}.

\noindent\textbf{Generalization Ability.} It is worth emphasizing that RigFace demonstrates strong generalization to out-of-domain identity images with unseen visual styles, maintaining effective control over appearance, even without any fine-tuning on the target domain. We observe that most existing methods are trained exclusively on real human face datasets. Therefore, to further evaluate the generalization ability of our model, we conduct the comparison on a broader set of character images, including various artistic and stylized identities that are significantly different from the training distribution. We collected a total of 500 images from \textit{pinterest} \footnote{https://fi.pinterest.com/} and \textit{unsplash} \footnote{https://unsplash.com/} websites to construct this dataset, which covers a wide range of face styles including comic, animation, oil painting, cyberpunk, and more. The comparison results in Fig. \ref{fig:comp1} and Tab. \ref{tab:qua_out} demonstrate that RigFace generalizes surprisingly well to these novel character images, even though it has never been trained on such data. Moreover, as shown in Tab. \ref{tab:out}, fine-tuning on specific datasets can further improve generation quality, enhancing both identity consistency and editability on top of the model’s zero-shot capability. However, the improvement is not substantial, suggesting that the model already achieves strong generalization performance after training on the current dataset.

\subsection{Ablation Study}
\label{sec:abl}
\noindent\textbf{Disentangled and pixel-aligned image condition.} To demonstrate the effectiveness of using disentangled conditions and of using image conditions for pose and lighting conditioning, we explore alternative designs, including 1) We don't use image condition. Instead, we concatenate the expression, pose, and lighting coefficients together into a vector to make the conditions entangled. Then we embed this vector through a single linear layer and then add the output into the time embedding of the Denoising UNet (\textbf{\textit{No-disent.}}). 2) We still use the coefficients instead of rendered images for pose and lighting condition, and we separate these coefficients to deal with them in different linear layers respectively. After that, we concatenate the ouputs from those linear layers into a vector and add it to the time embedding (\textbf{\textit{Coef.-sep.}}). As shown in Fig. \ref{fig:abl}, visualizations illustrate that decoupled conditions outperform entangled conditions. Soleing relying on 3DMM coefficients as conditions does not work on pose and lighting editing, which can also cause pose variation when it is not intended to edit. Quantitative results presented in Tab. \ref{tab:abl1} also show that facial expression editing benefits more from using expression coefficients as control conditions, while lighting editing performs better when using rendered images as guidance.

\noindent\textbf{Identity Encoder.} To demonstrate the effectiveness of Indentity encoder design, we conduct experiments 1) Replacing the Identity Encoder with ControlNet \cite{zhang2023adding} (\textbf{\textit{Control.}}); 2) Replacing the Identity Encoder with CLIP \cite{radford2021learning} image encoder (\textbf{\textit{CLIP-ID}}); 3) Replacing the Identity Encoder with a trainable Conv layer (\textbf{\textit{Conv-ID}}); 4) Using CLIP image encoder \cite{radford2021learning} to embed lignting conditions (\textbf{\textit{CLIP-light}}). As shown in Fig. \ref{fig:abl2}, ControlNet model will largely change the appearance of the input images and even fails on editing expressions. Using CLIP features as identity image features can preserve image similarity but fails to fully transfer details. Moreover, CLIP is unable to extract lighting features from rendering conditions, making the light editing in the fourth row less effective. A trainable Conv layer performs much better than CLIP in transfering facial details, but its performance in editing lighting and head pose is not effective enough. We speculate that this may be due to the identity condition and editing condition being placed together in the first layer of the UNet, increasing the burden on the first layer. Quantitative results are presented in Tab. \ref{tab:abl1} likewise, which is coherent with our qualitative observations.

\begin{figure}[t]
    \centering
    \includegraphics[width=0.48\textwidth]{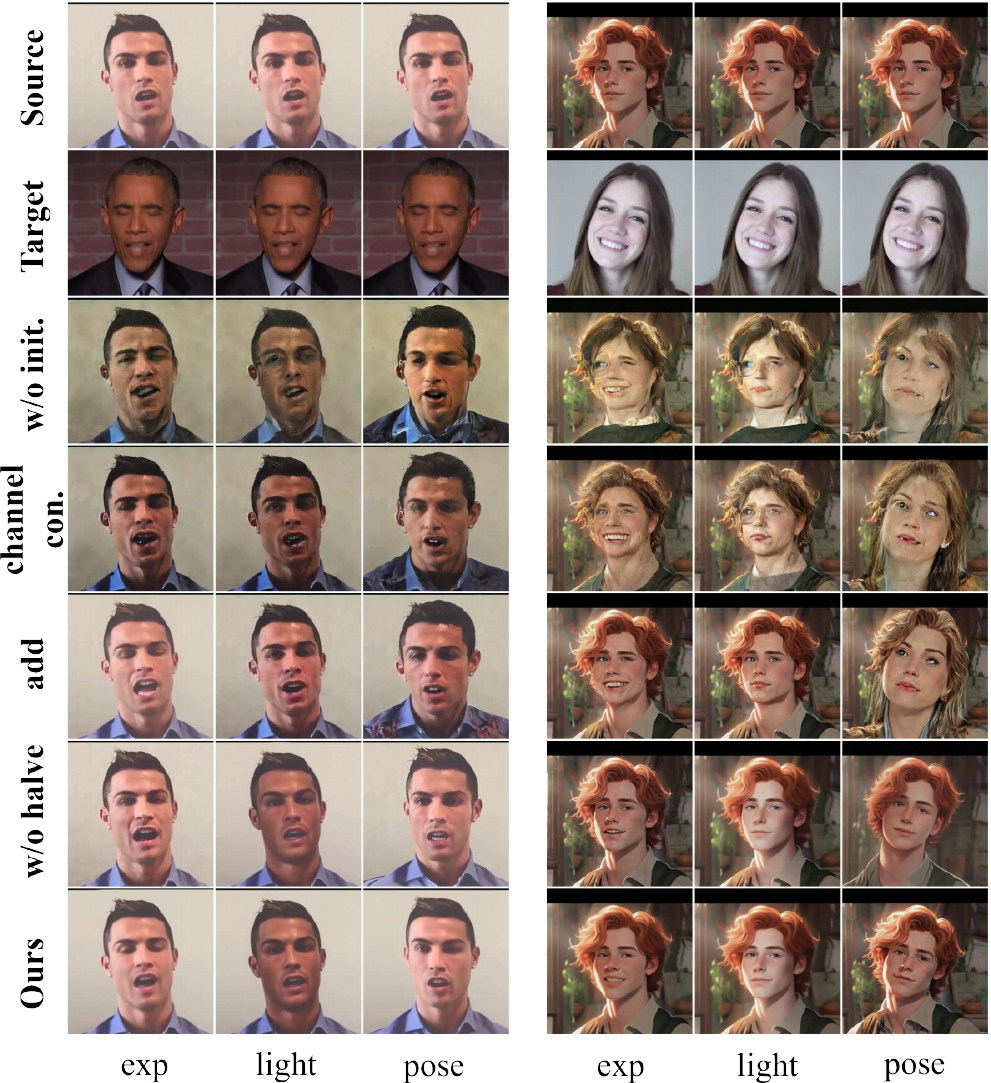}  
    \caption{\textbf{Ablation study on FaceFusion with Pretrained SD Initialization.} \textit{w/o init.} denotes training without SD parameter initialization; \textit{channel con.}, \textit{add}, and \textit{w/o halve} refer to different fusion strategies: channel-wise concatenation, element-wise addition, and spatial concatenation without truncating tokens after self-attention, respectively.}
    \label{fig:abl1}
\end{figure}

\begin{figure}[htbp]
    \centering
    \includegraphics[width=0.48\textwidth]{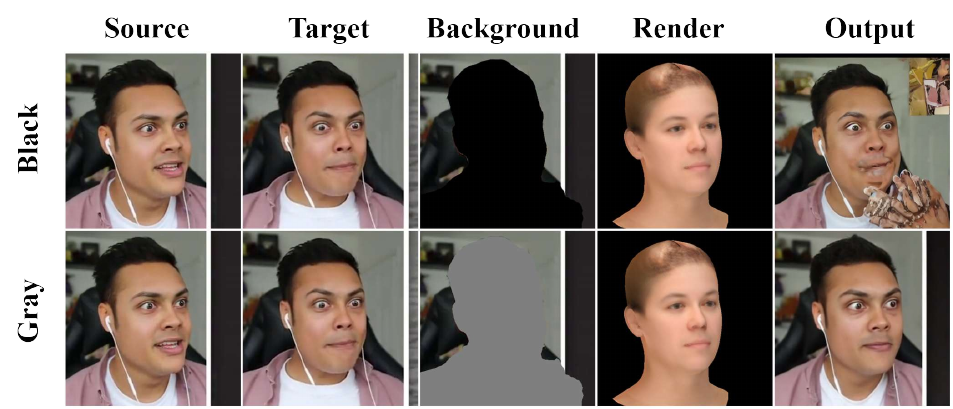}
    \caption{\textbf{Ablation study for different colors used to mask for the background image.} Black foreground is easy to clash with the color of the original images.}
    \label{fig:color}
\end{figure}

\noindent\textbf{FaceFusion with Pretrained SD Initialization.} To demonstrate the effectiveness of the proposed FaceFusion module, we investigate several strategies for integrating the self-attention features from the Identity Encoder into the Denoising UNet. Specifically, we compare: (1) element-wise addition of the identity and denoising features (\textbf{\textit{add}}); (2) channel-wise concatenation, where identity and denoising features are concatenated along the channel dimension before being passed through the self-attention layers (\textbf{\textit{channel con.}}); (3) concatenation along the spatial dimension, where the combined features are passed through the self-attention block and then directly sent to the cross-attention layer without truncation. This strategy increases the spatial size and necessitates introducing new cross-attention parameters (\textbf{\textit{w/o halve}}). On the other hand, to demonstrate the effectiveness of initializing our model with the pre-trained parameters of SD rather than random initialization, we conducted additional experiments using randomly initialized weights (\textbf{\textit{w/o init.}}). Qualitative results are shown in Fig. \ref{fig:abl1}. 

Among the feature fusion strategies between the identity encoder and the generative UNet, spatial concatenation yielded the best visual quality. In contrast, element-wise addition resulted in suboptimal identity preservation, particularly under pose editing conditions, while channel-wise concatenation failed to retain fine-grained facial identity details. We hypothesize that channel-wise concatenation merges identity and generative information at the same spatial locations, thereby blurring their respective roles and impairing the model’s ability to capture spatially aligned features. On the other hand, the results also show that random initialization leads to significantly worse performance under the same training budget. We attribute this to the large model size, limited dataset scale, and the complexity of editing conditions, which together hinder the model's capacity to learn effectively from scratch.

\noindent\textbf{Foreground Color.} We find that the color of the masked region has a significant impact on the experimental performance. As demonstrated in Fig. \ref{fig:color}, a black foreground can easily blend with a person's hair, clothing, or even the original image background, leading to errors in the model's background identification. This not only affects the preservation of the source image background but also impacts the final editing of the person. To solve this problem, we use gray to mask the human region, as it is a color that rarely clashes with the image background or the person.

\begin{figure*}[htbp]
    \centering
    \includegraphics[width=0.98\textwidth]{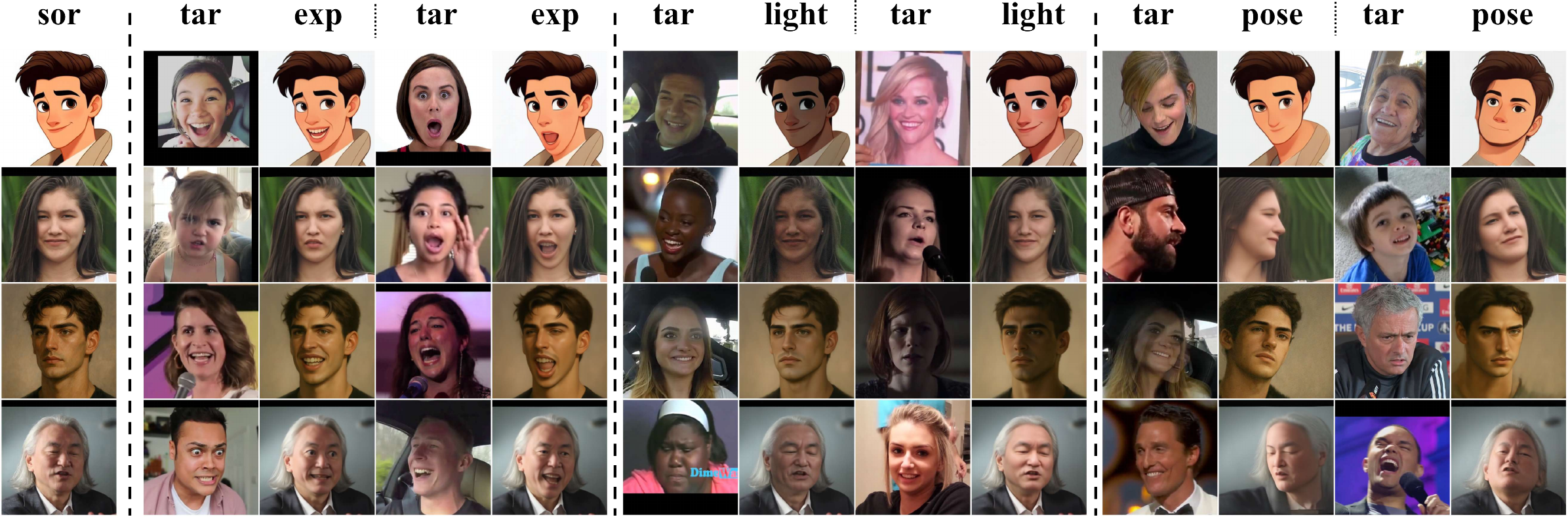}  

    \caption{\textbf{Illustration of extreme editing cases evaluated by our model,} including exaggerated expressions (e.g., wide-open mouth, frowning), challenging lighting conditions (e.g., dim illumination, strong shadows), and extreme head poses (e.g., upward tilt, full side profiles). \textit{sor} represents source image, \textit{tar} represents target image exhibiting the intended attribute, and \textit{exp, light, pose} represent the images generated by RigFace. The effects can be more clearly observed when zoomed in.}

    \label{fig:sp1}
\end{figure*}

\noindent \hypertarget{extreme}{\textbf{Extreme Conditions.}} We provide additional qualitative examples to illustrate the behavior of RigFace under extreme editing conditions, including expression, pose, and lighting. For each condition, two representative target images exhibiting extreme variations are used with different source identities to test the model's robustness. These extreme cases include exaggerated expressions (e.g., wide-open mouth, frowning), challenging lighting conditions (e.g., dim illumination, strong directional shadows), and large head pose deviations (e.g., upward tilt, full side profiles). As shown in Fig. \ref{fig:sp1}, RigFace maintains strong editing performance under extreme expression and lighting changes, preserving identity and ensuring visual coherence. However, for extreme pose variations, especially when there is a large spatial displacement between the source and target poses, the editing quality tends to degrade. In such cases, noticeable facial deformation may occur, as observed in the second and fourth examples in the pose editing column of Fig. \ref{fig:sp1}. This behavior is attributed to the need for significant background inpainting in extreme pose settings, which is a capability that lies outside the primary design scope of RigFace. This limitation has already been discussed in the main manuscript. The examples presented in Fig. \ref{fig:sp1} provide further insight into how the model performs under such challenging scenarios.

\begin{figure}[htbp]
\centering
\includegraphics[width=0.95\linewidth]{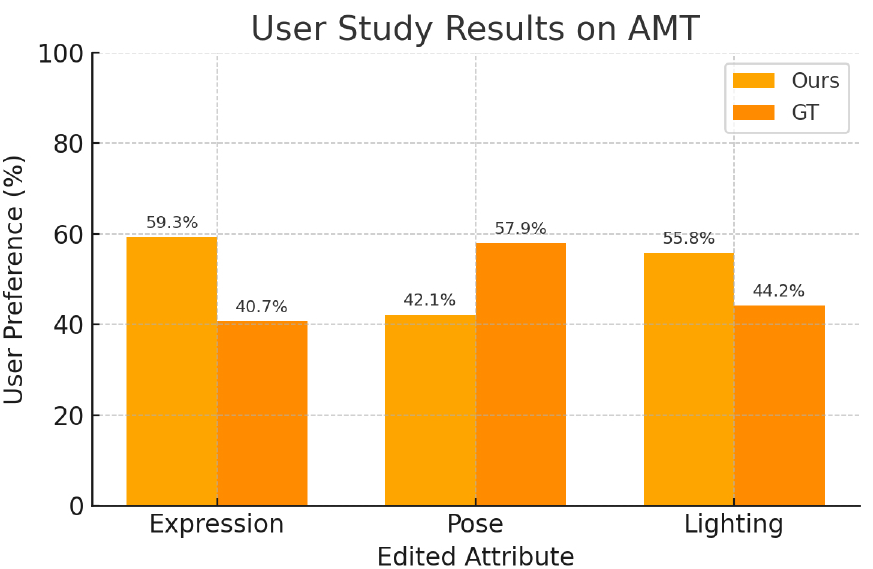}
\caption{\textbf{Results of user study} on the perceptual realism of expression, pose, and lighting edits.}
\label{fig:user}
\end{figure}

\noindent \hypertarget{user}{\textbf{User Study.}} To evaluate the perceptual realism of our generated images, we conducted a user study using Amazon Mechanical Turk (AMT). Each participant was presented with a pair of images, where one generated by our method and the other a real image. Each participant was asked to identify which image looked more realistic in terms of expression, pose, or lighting, depending on the specific editing type.

We collected 100 responses for each editing category. As shown in Fig. \ref{fig:user}, our method was selected as more realistic than the ground truth image in 59.3\% of expression edits, 42.1\% of pose edits, and 55.8\% of lighting edits, demonstrating that users found the generated images highly photorealistic and often indistinguishable from real photographs. These results confirm that our approach is capable of producing high-quality, perceptually convincing facial edits across various attributes.

\noindent \textbf{Limitations.} RigFace may struggle to preserve the original background when editing images with dramatic pose variations. The background inconsistency mainly arises from the need for extensive inpainting when the source and target poses differ significantly, which can lead to large spatial displacements. We provide a detailed discussion of this issue in the \hyperlink{extreme}{\textit{Extreme Conditions}} section. Furthermore, training two Stable Diffusion models simultaneously demands substantial computational resources.

\noindent \textbf{Future Work.} We plan to explore the integration of advanced inpainting models to improve background consistency under large pose variations, which could enhance the robustness of RigFace in more complex editing scenarios of head pose. In addition, we aim to investigate model distillation techniques to reduce inference overhead and improve scalability for real-world deployment.

\section{Conclusion}

We present RigFace, a latent diffusion model for facial appearance editing that surpasses prior GAN- and diffusion-based methods. It introduces a Spatial Attribute Provider for disentangled condition control and an Identity Encoder to preserve fine-grained identity features. By fully fine-tuning the pre-trained Stable Diffusion model, RigFace better adapts to the specific task of identity-preserving face editing. It enables precise manipulation of pose, expression, and lighting, with extensive experiments validating its effectiveness and high visual fidelity.



\bibliographystyle{IEEEtran}
\bibliography{tcsvt_references}

\end{document}